\documentclass{uber}

\definecolor{gtgreen}{HTML}{39FF14}
\definecolor{predorange}{HTML}{FFB300}

\renewcommand\contributionformat[2][]{{\small $#1$\,#2}}

\title{VLGA: Vision-Language-Geometry-Action Models for Autonomous Driving}

\author{Jin Yao$^{1,2}$}
\author{Dhruva Dixith Kurra$^{1}$}
\author{Tom Lampo$^{1}$}
\author{Zezhou Cheng$^{2\dagger}$} 
\author{Danhua Guo$^{1}$}
\author{Burhan Yaman$^{1\dagger\ddagger}$}
\affiliation[1]{Uber AV Labs}
\affiliation[2]{University of Virginia}
\contribution[\dagger]{Corresponding authors}\contribution[\ddagger]{Project Lead}

\project{\url{https://yaojin17.github.io/VLGA}}
\date{\today}
\renewcommand{\uberlogo}{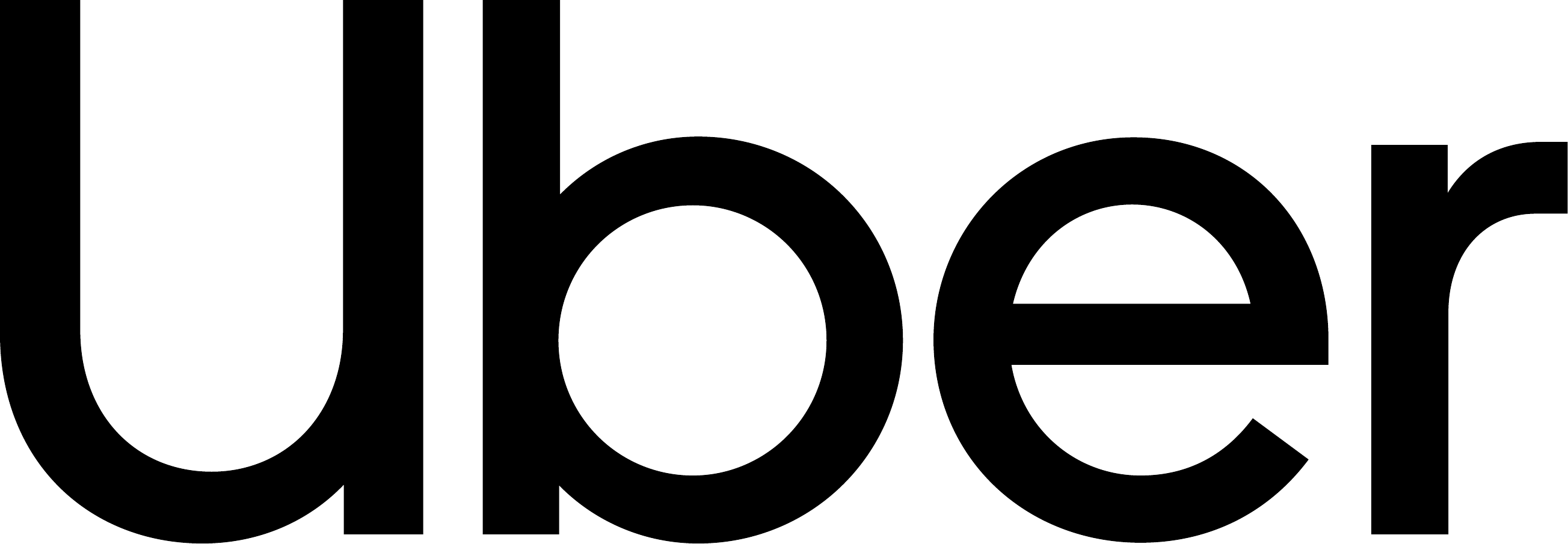}
\renewcommand{\uberlogosecond}{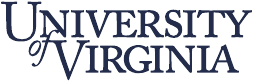}

\abstract{%
Vision-language-action (VLA) models can describe scenes and reason about them
in language, yet still struggle to ground their actions in the dense 3D world
around them. Existing approaches either inject features from a frozen 3D
foundation model without an objective that ensures the policy uses them, or
constrain geometry with sparse box and map losses that provide no dense spatial
signal. We introduce VLGA, the first vision-language-action model supervised to
reconstruct the dense 3D world it drives through. VLGA introduces geometry as a
fourth modality alongside vision, language, and action through a dedicated
expert supervised by a per-pixel pointmap regression loss against LiDAR.
Extensive experiments conducted on challenging nuScenes and Bench2Drive datasets
for open-loop and closed-loop evaluations, respectively, show the superiority of
VLGA over counterpart VLA methods. In particular, on open-loop nuScenes, VLGA
sets a new state of the art among VLA methods without ego status, with the
lowest L2 (0.50\,m average) and 3-second collision rate (0.18\%). On closed-loop
Bench2Drive, VLGA attains the state-of-the-art driving score of 79.08, +0.71
over the strongest prior VLA, at comparable efficiency and comfort.%
}

\begin{document}
\maketitle

\section{Introduction}
\label{sec:intro}

End-to-end vision-language-action (VLA) models for autonomous driving
inherit broad scene understanding and reasoning capabilities from pretrained
vision-language models~\cite{hwang2024emma,fu2025orion,tian2024drivevlm}, but how to ground such policies in
dense 3D structure remains an open question. Trajectory planning is
inherently a spatial task~\cite{hu2022st,jiang2023vad,hu2023planning}, and language reasoning alone cannot produce
the continuous spatial precision that safe driving demands.

\begin{figure*}[t]
\centering

\includegraphics[width=\linewidth]{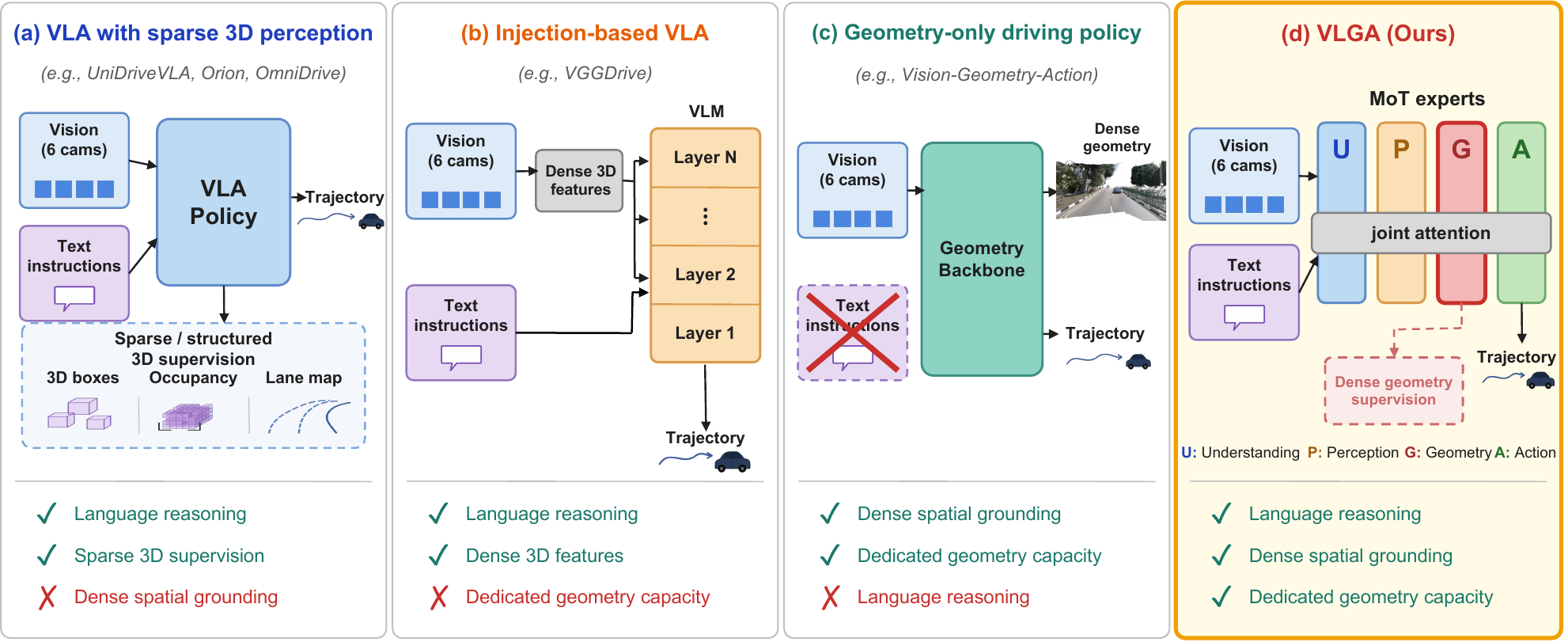}

\caption{
\textbf{Paradigms for grounding driving policies in 3D geometry.}
Existing approaches expose 3D structure in different ways, but each misses one key capability.
(a) VLAs with sparse 3D perception use structured 3D supervision such as boxes, occupancy, and lane maps, but lack dense spatial grounding.
(b) Injection-based VLAs expose dense 3D features to the language model, but lack dedicated geometry capacity.
(c) Geometry-only driving policies provide dense spatial grounding with dedicated geometry capacity, but remove language reasoning.
(d) \textbf{VLGA} preserves all three by introducing a parameter-isolated geometry expert supervised with dense geometry reconstruction.
}
\label{fig:paradigm}
\end{figure*}

Existing approaches expose 3D structure to a VLA policy in one of
three ways (Fig.~\ref{fig:paradigm}). The first relies on \textit{sparse
spatial perception}~\cite{li2026unidrivevla,fu2025orion,wang2025omnidrive}: a perception expert produces
query-decoded outputs such as 3D boxes, lane lines, and voxel
occupancy, which the action expert reads as a small set of discrete
predictions. The second \textit{injects dense features from a 3D
foundation model}~\cite{dust3r,wang2025vggt,pi3,dvgt2} into the language stream, either via
cross-attention at every LLM decoder layer~\cite{wang2026vggdrive} or as upstream
input tokens through BEV encoders~\cite{zhou2025opendrivevla} or 3D Q-Formers~\cite{wang2025omnidrive}, so
that the same LLM parameters process both language and 3D. The third,
exemplified by the recent \textit{Vision-Geometry-Action} (VGA)
framework~\cite{dvgt2}, dedicates the architecture to dense spatial
processing through a per-pixel pointmap reconstruction objective, but
removes the language stream entirely.

% =================================================================
%  Figure 1 (teaser): 4-panel paradigm comparison
% =================================================================

An effective VLA driving policy must bring together three capabilities:
\emph{language reasoning} to interpret intent and context,
\emph{dense spatial grounding} to perceive the continuous 3D
structure of the scene, and a \emph{dedicated geometry capacity} that
processes this structure using its own parameters rather than
the language model's. Yet no existing paradigm achieves all three
(Fig.~\ref{fig:paradigm}): sparse perception lacks dense grounding,
injection leaves no parameter budget dedicated to geometry, and VGA
drops language reasoning. We ask whether dense 3D
geometry can enter a VLA policy with a parameter budget of its own
while the language stream stays intact, letting a single policy keep
all three.

We close this gap with \textbf{VLGA}, a four-expert
vision-language-geometry-action driving policy. VLGA introduces 3D
geometry as a dedicated, parameter-isolated expert in a
Mixture-of-Transformers~\cite{liang2024mixture} backbone, alongside
the existing vision-language and perception experts. The perception
expert provides structured, instance-level cues, including surrounding
agents, lane geometry, and occupancy that the policy plans over,
whereas the geometry expert contributes the continuous per-pixel 3D
structure required for the fine spatial precision that planning
demands. Finally, the action expert conditions its predicted trajectory on
this geometry stream.
The geometry stream is supervised during training by a dense
per-pixel pointmap reconstruction loss against LiDAR, providing an
explicit training signal on the stream's 3D content rather than
relying on the action loss alone. Where prior vision-language-action
driving policies see, read, and act, VLGA also reconstructs.

Extensive experiments on open-loop and closed-loop settings show the exceptional capabilities of VLGA for enhanced driving. In particular, on open-loop nuScenes without ego status, VLGA ranks
first among VLA methods on 15 of the 16 planning metrics, including the
\textbf{lowest} L2 average (\textbf{0.50\,m}) and 3-second collision
rate (\textbf{0.18\%}), improving long-horizon safety over the
strongest prior VLA. On closed-loop
Bench2Drive~\cite{jia2024bench2drive}, VLGA attains a state-of-the-art Driving Score of
\textbf{79.08}, exceeding the strongest prior VLA by \textbf{+0.71}
at comparable efficiency and improved comfort. We summarize our contributions
as follows:
\begin{itemize}
    \item We propose VLGA, a Vision Language Geometry Action model, which introduces a dedicated geometry modality stream within
      a vision-language-action driving policy, supervised by a dense
      per-pixel pointmap reconstruction objective against LiDAR.
    \item Through extensive experiments, we show that this combination concentrates empirical gains
      on safety-critical metrics, reducing long-horizon collision
      over the strongest prior VLA on open-loop nuScenes and leading
      on spatial-precision-demanding closed-loop scenarios.
    \item On closed-loop Bench2Drive, the same dense geometric
      supervision delivers a new state-of-the-art driving score,
      demonstrating that the open-loop safety pattern replicates
      under closed-loop control.
\end{itemize}

\section{Related Work}
\label{sec:related_work}

\textbf{Vision-Language-Action Models for Autonomous Driving.}
Recent work has brought Vision-Language Models (VLMs) into autonomous driving to leverage their world knowledge and reasoning for long-tail scenario handling~\cite{hwang2024emma,fu2025orion,zeng2025futuresightdrive,zhou2025opendrivevla,yang2025drivemoe,wang2025alpamayo,li2025recogdrive,wang2025omnidrive}. Dual-system approaches~\cite{tian2024drivevlm,jiang2024senna} pair a slow VLM with a fast end-to-end driving model, while more recent single-system Vision-Language-Action (VLA) architectures emit the trajectory directly, either as language tokens~\cite{hwang2024emma,wang2025omnidrive,zhang2024wisead,xing2025openemma,chi2025impromptu} or through a coupled action decoder~\cite{zhou2025opendrivevla,zhou2025autovla,yang2025drivemoe,fu2025orion,li2025recogdrive,zeng2025futuresightdrive,renz2025simlingo,renz2024carllava,jiang2025diffvla,li2025drivevla,wang2025alpamayo}. Reinforcement-learning fine-tuning has also been explored to align trajectories with task rewards~\cite{zhou2025autovla,li2025recogdrive,jiang2025alphadrive,luo2025adathinkdrive}. Despite this diversity, dense 3D scene understanding remains underdeveloped: these policies encode scene semantics far more strongly than per-pixel 3D structure. In contrast, our VLGA introduces 3D geometry as a dedicated supervised modality stream alongside vision, language, and perception.

\textbf{Geometry-Aware Driving Policies.}
Two existing patterns bring 3D understanding into driving policies. \textit{Injection-based} approaches feed features from a pretrained 3D foundation model~\cite{dust3r,mast3r,wang2025vggt,pi3,dvgt2} into the LLM's hidden state, whether by cross-attention, BEV tokens, or 3D Q-Formers~\cite{wang2026vggdrive,zhou2025opendrivevla,wang2025omnidrive}; one set of LLM parameters must then serve both language and 3D, with no objective ensuring the injected geometry is actually used. A second, \textit{geometry-only} pattern, exemplified by the Vision-Geometry-Action framework of DVGT-2~\cite{dvgt2}, dedicates the architecture to dense pointmap reconstruction but drops the language stream. VLGA instead allocates dedicated, parameter-isolated capacity to 3D geometry within a Mixture-of-Transformers~\cite{liang2024mixture} and supervises it with a dense per-pixel pointmap objective, keeping the language stream intact.

\textbf{Sparse and Dense 3D Representations for End-to-End Driving.}
A complementary question is at what granularity 3D structure should enter the policy. Sparse 3D queries from object-query detectors~\cite{wang2022detr3d,liu2022petr,lin2022sparse4d,liu2023sparsebev} and their end-to-end planning extensions~\cite{sun2025sparsedrive,zhu2025sparsead} represent the scene compactly through object-level queries, a paradigm that UniDriveVLA~\cite{li2026unidrivevla} brought into the VLA framework. This object-level granularity suffices for high-level decisions such as which agents to yield to or which gaps to enter, but is coarse for safety-critical scenarios such as maintaining tight lateral clearance from a parked vehicle or anticipating the swept volume of an oncoming vehicle, both of which require continuous per-pixel 3D understanding rather than a list of discrete object boxes. In contrast, our VLGA combines sparse and dense perception as complementary streams, with empirical gains concentrated on the safety-critical scenarios that demand dense spatial precision (Sec.~\ref{sec:results}).

\section{Methodology}
\label{sec:methodology}

% =================================================================
%  Figure 2: VLGA architecture (placement is at top of Method section)
% =================================================================
\begin{figure*}[t]
\centering
\includegraphics[width=\linewidth]{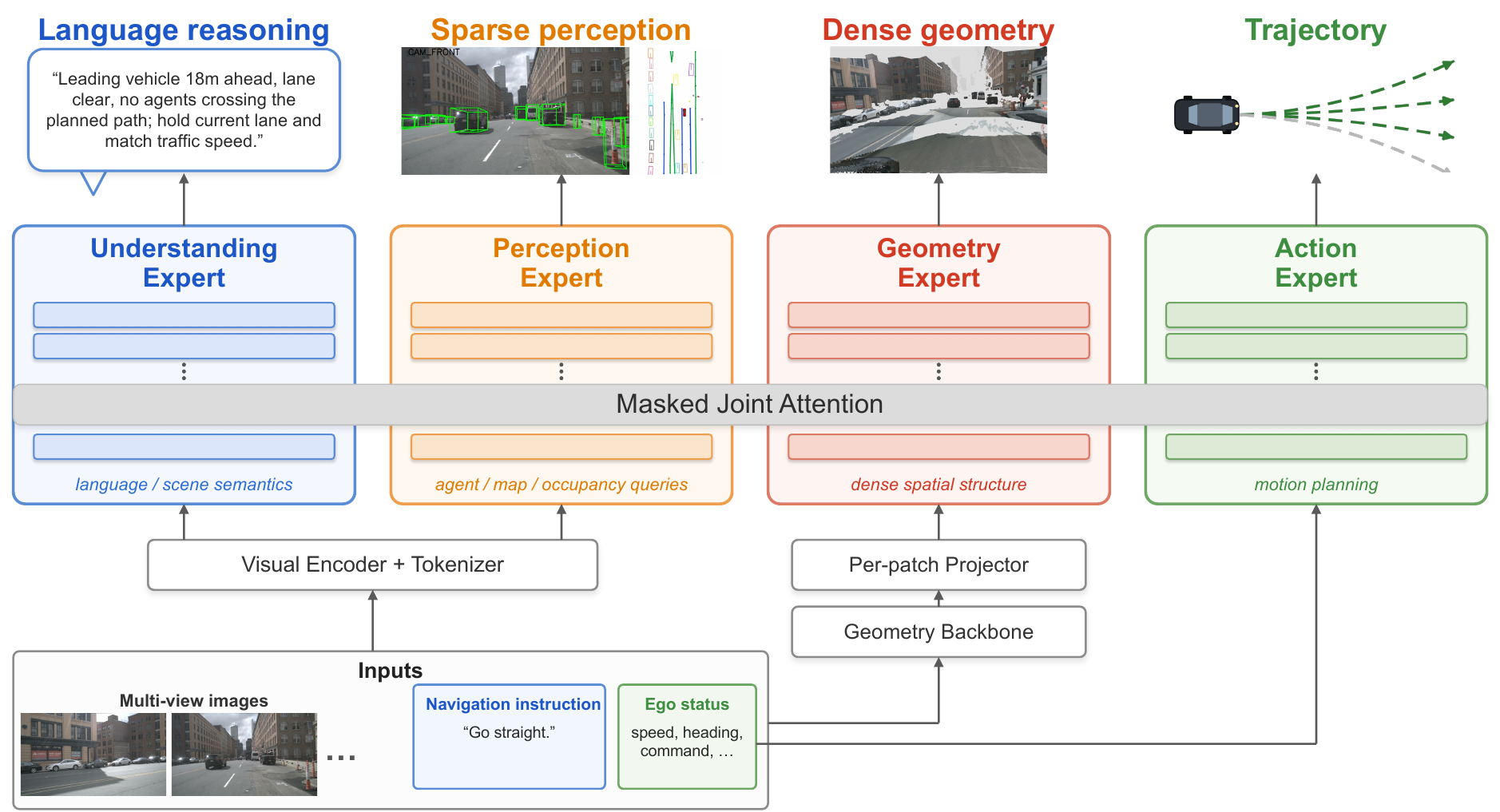}

\caption{\textbf{VLGA architecture.} A four-expert Mixture-of-Transformers
coupled by masked joint attention: an understanding expert ($U$,
language/scene semantics), a perception expert ($P$, sparse agent/map/occupancy
queries), our new geometry expert ($G$, dense spatial structure), and an action
expert ($A$, motion planning). $A$ attends to $U$, $P$, $G$ and conditions on
ego status to emit the trajectory.}
\label{fig:architecture}
\end{figure*}

We introduce \textbf{VLGA}, a vision-language-action driving policy
whose geometry stream is supervised by dense pointmap reconstruction
(Fig.~\ref{fig:architecture}). We first formalize the planning task
and notation (Sec.~\ref{sec:method:prelim}) and overview the
four-expert architecture (Sec.~\ref{sec:method:overview}). We then
detail the geometry expert
(Sec.~\ref{sec:method:geo_expert}), the dense pointmap
reconstruction objective that supervises this stream
(Sec.~\ref{sec:method:pointmap}), and the two-stage training
schedule that integrates it with the action expert
(Sec.~\ref{sec:method:training}).

\subsection{Preliminaries}
\label{sec:method:prelim}

Given multi-view camera observations $\mathcal{I} \in \mathbb{R}^{V \times H \times W \times 3}$ from $V$ surround cameras, an ego status $\mathbf{s}$, and a high-level navigation instruction $\ell$, an end-to-end driving policy $\pi$ predicts the future trajectory $\mathbf{T} = \{(x_t, y_t)\}_{t=1}^{T_f}$ of the ego vehicle over a planning horizon of $T_f$ steps:
\begin{equation}
    \mathbf{T} = \pi(\mathcal{I}, \mathbf{s}, \ell).
\end{equation}
Recent vision-language-action (VLA) models implement $\pi$ as a Mixture-of-Transformers (MoT)~\cite{liang2024mixture} hosting modality-specialized experts that typically include a vision-language expert (the VLM itself), a perception expert, and an action expert. We extend the MoT paradigm by adding a fourth expert dedicated to geometry. Throughout this paper, $U$, $P$, $G$, $A$ denote the understanding (vision-language), perception, geometry, and action experts of the MoT.

\subsection{Architecture overview}
\label{sec:method:overview}

The VLGA model consists of a multi-view vision encoder, a language tokenizer, a four-expert MoT block, and the geometry backbone. Within the MoT block, the new geometry expert is added alongside the vision-language (VLM), perception, and action experts inherited from prior VLAs. The vision-language backbone, perception expert, and geometry backbone all retain weights from their respective pretrained checkpoints (specified in Sec.~\ref{sec:setup}) and remain frozen throughout training. The trainable components introduced in this paper are the geometry expert, a per-patch geometry projector, the action expert, and a lightweight dense pointmap decoder used to supervise the geometry stream during training. Fig.~\ref{fig:architecture} illustrates the overall architecture.

Concretely, each MoT layer couples the experts through masked joint attention~\cite{liang2024mixture}: per-expert Q/K/V projections are concatenated across $\{U, P, G, A\}$ and attend under a visibility mask $\mathbf{M}$, followed by per-expert output projection and feed-forward layers. The mask $\mathbf{M}$ retains the $U$/$P$/$A$ attention pattern from UniDriveVLA~\cite{li2026unidrivevla} and integrates the geometry stream into it: the geometry tokens attend to the understanding and perception tokens, and the action expert additionally attends to the geometry tokens.

\textbf{Ego-status conditioning.} Following UniDriveVLA~\cite{li2026unidrivevla}, the perception expert $P$ includes a self-prediction head for ego-status $\mathbf{s}$ (linear velocity, acceleration, and a one-hot high-level driving command). At inference, the action expert can be conditioned on either ground-truth ego-status, $\mathbf{s} = \mathbf{s}^{\text{gt}}$ (the \textit{with-ego-status} protocol), or on the perception expert's self-prediction, $\mathbf{s} = \hat{\mathbf{s}}$ (the \textit{without-ego-status} protocol). The two protocols use the same model weights and differ only in the eval-time source of $\mathbf{s}$.

\subsection{Geometry expert from a pretrained geometry backbone}
\label{sec:method:geo_expert}

VLGA reuses the geometry backbone~\cite{dvgt2} as the source of the geometry stream rather than learning dense geometry from scratch. The geometry backbone operates on the same multi-view camera input as the vision-language backbone, at resolution $960 \times 544$. For each of the $V = 6$ cameras, it emits a $60 \times 34 = 2{,}040$ grid of per-patch features at hidden dimension $d_g$, giving $V \times 2{,}040 = 12{,}240$ geometry tokens per scene. We keep the full per-patch grid rather than pooling, since the dense reconstruction objective (Sec.~\ref{sec:method:pointmap}) needs per-pixel spatial resolution. A lightweight per-patch projector $f_\text{proj}: \mathbb{R}^{d_g} \to \mathbb{R}^{d_\text{MoT}}$ maps each geometry token into the MoT token space:
\begin{equation}
    \mathbf{g}_i = f_\text{proj}(\mathbf{f}_i^\text{geo}), \quad i \in \{1, \ldots, 12{,}240\}.
\end{equation}
These projected tokens form the geometry stream and enter the MoT as the $G$ expert's input, interacting with the other three experts via masked joint attention. The action expert attends to $\{U, P, G\}$ and produces the trajectory via flow matching~\cite{lipman2022flow}.

\subsection{Dense pointmap supervision}
\label{sec:method:pointmap}

To provide direct supervision on the geometry stream's 3D content rather than relying on the action loss alone to shape what the stream encodes, we add a dense per-pixel pointmap reconstruction objective. A lightweight five-layer transformer decoder~\cite{vaswani2017attention} $\mathcal{D}$ consumes the geometry tokens $\{\mathbf{g}_i\}$ and produces, for each pixel patch $p$, a predicted 3D point $\hat{\mathbf{x}}_p \in \mathbb{R}^3$ in the ego LiDAR frame together with an uncertainty logit $c_p \in \mathbb{R}$:
\begin{equation}
    (\hat{\mathbf{x}}_p, c_p) = \mathcal{D}(\{\mathbf{g}_i\})_p, \quad p \in \{1, \ldots, P_\text{tot}\},
\end{equation}
where $P_\text{tot}$ is the total number of patches across all cameras. We target points in the ego frame rather than camera-frame depth, so the geometry stream lives in the same coordinate system the policy plans in. Following a Pi3-style confidence-weighted regression objective~\cite{pi3,kendall2017uncertainties}, the per-patch loss is
\begin{equation}
    \mathcal{L}_\text{pmap} = \frac{1}{|\mathcal{P}|} \sum_{p \in \mathcal{P}} \left( \frac{\lVert \hat{\mathbf{x}}_p - \mathbf{x}_p^\text{gt} \rVert_1}{b_p} + \log b_p \right), \quad b_p = \text{softplus}(c_p),
\end{equation}
where $\mathcal{P}$ is the set of patches whose ground-truth pointmap is valid (depth within $[0.5, 80.0]$ metres). Ground-truth pointmaps are derived by accumulating LiDAR sweeps and projecting them onto each camera. No LiDAR is required at inference: the decoder $\mathcal{D}$ is used only to compute the training signal and is discarded once training is complete.

\subsection{Two-stage training schedule}
\label{sec:method:training}

We adopt a two-stage schedule so that the randomly-initialized geometry components can first warm up under explicit pointmap supervision before being co-adapted with the action expert. This avoids interference between the new geometric loss signal and the inherited action expert's optimization trajectory. In the \textit{geometry stage}, the action expert and all inherited streams are frozen and only the geometric components (geometry expert, projector, decoder $\mathcal{D}$) are trained, supervised by $\mathcal{L}_\text{pmap}$ alone at unit weight. In the \textit{joint stage}, the action expert is unfrozen and the action loss $\mathcal{L}_\text{act}$ is added; $\mathcal{L}_\text{pmap}$ is retained at a reduced weight $\lambda_\text{pmap} = 0.1$ to avoid interfering with action training. The per-stage objectives are
\begin{align}
    \mathcal{L}_\text{geom} &= \mathcal{L}_\text{pmap}, \\
    \mathcal{L}_\text{joint} &= \mathcal{L}_\text{act} + \lambda_\text{pmap}\, \mathcal{L}_\text{pmap}.
\end{align}
Numerical training hyperparameters (learning rates, batch sizes, epoch counts, optimizer) are deferred to Sec.~\ref{sec:setup}.

\section{Experiments and Results}
\label{sec:results}

% =================================================================
%  Table 1: nuScenes open-loop planning (dual protocol: ST-P3 + UniAD)
%  Within each ego block: non-VLA methods on top, VLA below, VLGA (Ours) last.
% =================================================================
\begin{table*}[t]
\centering
\caption{Open-loop planning on the nuScenes validation split.
L2 displacement and collision rate are reported under the ST-P3 and
UniAD protocols, and lower is better throughout. A superscript
$^{\ast}$ marks methods that use ground-truth ego status,
$^{\ddagger}$ marks SparseDrive re-evaluated with the GPT-Driver
protocol, and $^{\dag}$ marks FSDrive trained with an extended schedule.}
\label{tab:nusc_planning}
\setlength{\tabcolsep}{3pt}
\footnotesize
\resizebox{\textwidth}{!}{
\begin{tabular}{lcccccccc|cccccccc|c}
\toprule
\multirow{4}{*}{Method}
 & \multicolumn{8}{c}{ST-P3 metrics}
 & \multicolumn{8}{c}{UniAD metrics}
 & \multirow{4}{*}{LLM} \\
\cmidrule(lr){2-9} \cmidrule(lr){10-17}
 & \multicolumn{4}{c}{L2 (m) $\downarrow$}
 & \multicolumn{4}{c}{Collision (\%) $\downarrow$}
 & \multicolumn{4}{c}{L2 (m) $\downarrow$}
 & \multicolumn{4}{c}{Collision (\%) $\downarrow$} & \\
\cmidrule(lr){2-5} \cmidrule(lr){6-9} \cmidrule(lr){10-13} \cmidrule(lr){14-17}
 & 1s & 2s & 3s & Avg.\ & 1s & 2s & 3s & Avg.\ & 1s & 2s & 3s & Avg.\ & 1s & 2s & 3s & Avg.\ & \\
\midrule
\multicolumn{18}{c}{\textit{Methods with Ego Status}} \\
\midrule
ST-P3$^{\ast}$~\citep{hu2022st}         & 1.33 & 2.11 & 2.90 & 2.11 & 0.23 & 0.62 & 1.27 & 0.71 & --   & --   & --   & --   & --   & --   & --   & --   & --            \\
VAD$^{\ast}$~\citep{jiang2023vad}           & 0.17 & 0.34 & 0.60 & 0.37 & 0.04 & 0.27 & 0.67 & 0.33 & --   & --   & --   & --   & --   & --   & --   & --   & --            \\
UniAD$^{\ast}$~\citep{hu2023planning}         & --   & --   & --   & --   & --   & --   & --   & --   & 0.20 & 0.42 & \textbf{0.75} & 0.46 & 0.02 & 0.25 & 0.84 & 0.37 & --            \\
BEV-Planner$^{\ast}$~\citep{li2024ego}   & 0.16 & 0.32 & 0.57 & 0.35 & \textbf{0.00} & 0.29 & 0.73 & 0.34 & --   & --   & --   & --   & --   & --   & --   & --   & --            \\
HPP$^{\ast}$~\citep{liu2025hybrid}           & 0.26 & 0.37 & 0.59 & 0.40 & 0.02 & \textbf{0.05} & \textbf{0.11} & \textbf{0.06} & 0.30 & 0.61 & 1.15 & 0.72 & 0.03 & 0.07 & \textbf{0.35} & \textbf{0.15} & --            \\
DVGT-2$^{\ast}$~\citep{dvgt2}        & 0.20 & 0.37 & 0.66 & 0.41 & 0.04 & 0.14 & 0.47 & 0.22 & --   & --   & --   & --   & --   & --   & --   & --   & --            \\
RDA-Driver$^{\ast}$~\citep{huang2024making}    & 0.17 & 0.37 & 0.69 & 0.40 & 0.01 & \textbf{0.05} & 0.26 & 0.10 & 0.23 & 0.73 & 1.54 & 0.80 & \textbf{0.00} & 0.13 & 0.83 & 0.32 & LLaVA-7B      \\
OmniDrive$^{\ast}$~\citep{wang2025omnidrive}     & \textbf{0.14} & 0.29 & 0.55 & 0.33 & \textbf{0.00} & 0.13 & 0.78 & 0.30 & --   & --   & --   & --   & --   & --   & --   & --   & LLaVA-7B      \\
Orion$^{\ast}$~\citep{fu2025orion}         & 0.17 & 0.31 & 0.55 & 0.34 & 0.05 & 0.25 & 0.80 & 0.37 & --   & --   & --   & --   & --   & --   & --   & --   & LLaVA-7B      \\
FSDrive$^{\ast\dag}$~\citep{zeng2025futuresightdrive}   & \textbf{0.14} & \textbf{0.25} & \textbf{0.46} & \textbf{0.28} & 0.03 & 0.06 & 0.21 & 0.10 & \textbf{0.18} & \textbf{0.39} & 0.77 & \textbf{0.45} & \textbf{0.00} & \textbf{0.06} & 0.42 & 0.16 & Qwen2-VL-3B   \\
AutoVLA$^{\ast}$~\citep{zhou2025autovla}       & 0.25 & 0.46 & 0.73 & 0.48 & 0.07 & 0.07 & 0.26 & 0.13 & 0.33 & 0.81 & 1.45 & 0.86 & 0.08 & 0.11 & 0.85 & 0.35 & Qwen2.5-VL-3B \\
OpenDriveVLA$^{\ast}$~\citep{zhou2025opendrivevla}  & \textbf{0.14} & 0.30 & 0.55 & 0.33 & 0.02 & 0.07 & 0.22 & 0.10 & 0.19 & 0.58 & 1.24 & 0.67 & 0.02 & 0.18 & 0.70 & 0.30 & Qwen2.5-VL-3B \\
UniDriveVLA-Base$^{\ast}$~\citep{li2026unidrivevla}  & 0.23 & 0.40 & 0.65 & 0.43 & 0.04 & 0.09 & 0.18 & 0.10 & 0.30 & 0.69 & 1.32 & 0.77 & 0.03 & 0.13 & 0.52 & 0.23 & Qwen3-VL-2B \\
UniDriveVLA-Large$^{\ast}$~\citep{li2026unidrivevla} & 0.24 & 0.40 & 0.63 & 0.42 & 0.03 & 0.09 & 0.16 & 0.10 & 0.30 & 0.66 & 1.25 & 0.74 & 0.02 & 0.18 & 0.40 & 0.20 & Qwen3-VL-8B \\
VGGDrive$^{\ast}$~\citep{wang2026vggdrive}      & \textbf{0.14} & 0.28 & 0.51 & 0.31 & 0.02 & 0.10 & 0.55 & 0.22 & --   & --   & --   & --   & --   & --   & --   & --   & Qwen2.5-VL-7B \\
\midrule
VLGA-Base (Ours)$^{\ast}$  & 0.21 & 0.39 & 0.64 & 0.41 & 0.03 & 0.07 & 0.19 & 0.10 & 0.28 & 0.67 & 1.31 & 0.75 & 0.02 & 0.12 & 0.60 & 0.24 & Qwen3-VL-2B \\
VLGA-Large (Ours)$^{\ast}$ & 0.22 & 0.39 & 0.62 & 0.41 & 0.02 & 0.07 & 0.16 & 0.08 & 0.29 & 0.65 & 1.27 & 0.74 & \textbf{0.00} & 0.15 & 0.38 & 0.18 & Qwen3-VL-8B \\
\midrule
\multicolumn{18}{c}{\textit{Methods without Ego Status}} \\
\midrule
VAD~\citep{jiang2023vad}                    & 0.41 & 0.70 & 1.05 & 0.72 & 0.03 & 0.19 & 0.43 & 0.21 & --   & --   & --   & --   & --   & --   & --   & --   & --           \\
UniAD~\citep{hu2023planning}                  & 0.45 & 0.70 & 1.04 & 0.73 & 0.62 & 0.58 & 0.63 & 0.61 & 0.59 & 1.01 & \textbf{1.48} & 1.03 & 0.16 & 0.51 & 1.64 & 0.77 & --           \\
OccWorld~\citep{zheng2024occworld}               & 0.39 & 0.73 & 1.18 & 0.77 & 0.11 & 0.19 & 0.67 & 0.32 & 0.52 & 1.27 & 2.41 & 1.40 & 0.12 & 0.40 & 2.08 & 0.87 & --           \\
BEV-Planner~\citep{li2024ego}            & 0.30 & 0.52 & 0.83 & 0.55 & 0.10 & 0.37 & 1.30 & 0.59 & --   & --   & --   & --   & --   & --   & --   & --   & --           \\
HPP~\citep{liu2025hybrid}                    & 0.41 & 0.61 & 0.86 & 0.63 & 0.03 & 0.08 & 0.24 & 0.12 & 0.48 & 0.91 & 1.54 & 0.97 & 0.03 & 0.17 & 0.68 & 0.29 & --           \\
SparseDrive$^{\ddagger}$~\citep{sun2025sparsedrive} & 0.28 & 0.53 & 0.84 & 0.55 & 0.03 & \textbf{0.05} & \textbf{0.15} & \textbf{0.08} & 0.38 & 0.92 & 1.66 & 0.99 & 0.02 & \textbf{0.08} & 0.53 & \textbf{0.21} & --           \\
ELM~\citep{zhou2024embodied}                    & --   & --   & --   & --   & --   & --   & --   & --   & \textbf{0.34} & 1.23 & 2.57 & 1.38 & 0.12 & 0.50 & 2.36 & 0.99 & BLIP2-2.7B   \\
OmniDrive~\citep{wang2025omnidrive}              & 0.40 & 0.80 & 1.32 & 0.84 & 0.04 & 0.46 & 2.32 & 0.94 & --   & --   & --   & --   & --   & --   & --   & --   & LLaVA-7B     \\
FSDrive~\citep{zeng2025futuresightdrive}                & 0.28 & 0.52 & 0.80 & 0.53 & 0.06 & 0.13 & 0.32 & 0.17 & 0.40 & 0.89 & 1.60 & 0.96 & 0.07 & 0.12 & 1.02 & 0.40 & Qwen2-VL-3B  \\
UniDriveVLA-Base~\citep{li2026unidrivevla}       & 0.28 & 0.51 & 0.82 & 0.54 & 0.08 & 0.13 & 0.31 & 0.17 & 0.37 & 0.89 & 1.62 & 0.96 & 0.08 & 0.27 & 0.88 & 0.41 & Qwen3-VL-2B \\
UniDriveVLA-Large~\citep{li2026unidrivevla}      & 0.27 & 0.49 & 0.77 & 0.51 & 0.03 & 0.10 & 0.21 & 0.11 & 0.36 & \textbf{0.83} & 1.50 & \textbf{0.90} & 0.02 & 0.23 & 0.55 & 0.27 & Qwen3-VL-8B \\
\midrule
VLGA-Base (Ours)       & \textbf{0.26} & 0.50 & 0.81 & 0.52 & 0.03 & 0.10 & 0.28 & 0.14 & 0.36 & 0.88 & 1.60 & 0.95 & 0.02 & 0.25 & 0.78 & 0.35 & Qwen3-VL-2B \\
VLGA-Large (Ours)      & \textbf{0.26} & \textbf{0.48} & \textbf{0.76} & \textbf{0.50} & \textbf{0.02} & 0.06 & 0.18 & 0.09 & \textbf{0.34} & \textbf{0.83} & 1.52 & \textbf{0.90} & \textbf{0.00} & 0.15 & \textbf{0.50} & 0.22 & Qwen3-VL-8B \\
\bottomrule
\end{tabular}}
\end{table*}

\subsection{Experimental Setup}
\label{sec:setup}

\textbf{Datasets.} We evaluate VLGA on two driving benchmarks. nuScenes
\cite{caesar2020nuscenes} provides 6019 validation samples from 150 urban driving scenes
captured by a six-camera rig, and we use it for open-loop trajectory
planning. Bench2Drive~\cite{jia2024bench2drive} provides 220 closed-loop driving routes
across 12 CARLA~\cite{dosovitskiy2017carla} simulator towns, scored by an official evaluation
harness that runs the policy under closed-loop control, and we use it
for closed-loop driving evaluation.

\textbf{Metrics.} On nuScenes we report L2 displacement and collision
rate at 1\,s, 2\,s, and 3\,s and their averages, under
ST-P3~\cite{hu2022st} and UniAD~\cite{hu2023planning} protocols. On
Bench2Drive we report the closed-loop metrics, Driving Score
(DS), Success Rate (SR), Efficiency, and Comfortness, plus per-skill SR
over the five categories of Merging, Overtaking, Emergency
Brake, Give Way, and Traffic Sign.

\textbf{Implementation Details.} VLGA comes in two variants,
VLGA-Base (Qwen3-VL-2B~\cite{qwen3vl}) and VLGA-Large (Qwen3-VL-8B),
with DVGT-2~\cite{dvgt2} as the geometry backbone. All frames are
resized to $960 \times 544$, and the vision-language backbone and
perception expert, initialized from UniDriveVLA~\cite{li2026unidrivevla},
are frozen throughout. We train on the nuScenes train split
(28{,}130 keyframes) and the Bench2Drive train routes, with 10/3 epochs
(nuScenes/Bench2Drive) in the geometry stage and 30/7 in the joint
stage (Sec.~\ref{sec:method:training}). Both stages use
AdamW~\cite{loshchilov2017decoupled} at base learning rate
$5 \times 10^{-5}$, effective batch size 128, and EMA
(momentum $2 \times 10^{-4}$, warmup 2000), on 8 H100 GPUs.

\subsection{Main Results}
\label{sec:main_results}

\textbf{Open-loop planning on nuScenes
(Tab.~\ref{tab:nusc_planning}).}
We focus on the \textit{without-ego-status} column, since with-ego
evaluation on nuScenes is dominated by ego-state leakage, where
kinematic extrapolation alone achieves strong L2 without scene
understanding~\cite{li2024ego}. In this leakage-free
setting, VLGA-Large ranks first among VLA methods on 15 of the 16 L2
and collision metrics, trailing only on UniAD L2 at 3\,s (1.52 vs.\
1.50). Its ST-P3 L2 is the lowest of all methods (\textbf{0.50}\,m
average), and its collision rate is the lowest among VLA methods at
every horizon (\textbf{0.18\%} at 3\,s). The same pattern holds at 2B scale: VLGA-Base
reduces the same-scale prior VLA's collision rate from 0.41\% to
\textbf{0.35\%}, indicating that the improvement stems from the
geometry stream rather than larger model capacity. The gains
concentrate on safety rather than mean-trajectory accuracy: even where
the contemporaneous DVGT-2~\cite{dvgt2} and VGGDrive~\cite{wang2026vggdrive}
report lower L2 (with ego status), VLGA's 3-second collision rate
remains roughly 2.5--3$\times$ lower.

\textbf{Closed-loop driving on Bench2Drive
(Tab.~\ref{tab:b2d_closed_loop} and
Tab.~\ref{tab:b2d_multi_ability}).}
VLGA achieves the highest Driving Score of all methods, \textbf{79.08}, beating the prior state of the art, UniDriveVLA, by \textbf{+0.71}. It improves on this baseline in Success Rate and Comfortness at comparable Efficiency, and raises the per-skill mean from 51.53 to \textbf{53.04\%}; Sec.~\ref{sec:analysis} examines which skills drive these gains.
% =================================================================
%  Table 3: Bench2Drive closed-loop main metrics
% =================================================================
\begin{table*}[t]
\centering
\caption{Closed-loop driving on Bench2Drive in CARLA. The leading
\textit{Avg.\ L2} column reports open-loop 3\,s trajectory error,
where lower is better; all other columns are official closed-loop
metrics, where higher is better.}
\label{tab:b2d_closed_loop}
\setlength{\tabcolsep}{6pt}
\small
\begin{tabular}{l|c|cccc}
\toprule
Method & Avg.\ L2 $\downarrow$ & Driving Score $\uparrow$ & Success Rate (\%) $\uparrow$ & Efficiency $\uparrow$ & Comfortness $\uparrow$ \\
\midrule
AD-MLP~\citep{zhai2023rethinking}            & 3.64 & 18.05 &  0.00 &  48.45 & 22.63 \\
VAD~\citep{jiang2023vad}               & 0.91 & 42.35 & 15.00 & 157.94 & 46.01 \\
SparseDrive~\citep{sun2025sparsedrive}       & 0.87 & 44.54 & 16.71 & 170.21 & 48.63 \\
GenAD~\citep{zheng2024genad}             & --   & 44.81 & 15.90 & --     & --    \\
UniAD~\citep{hu2023planning}             & 0.73 & 45.81 & 16.36 & 129.21 & 43.58 \\
MomAD~\citep{song2025don}             & 0.82 & 47.91 & 18.11 & 174.91 & \textbf{51.20} \\
SeerDrive~\citep{zhang2025future}         & 0.66 & 58.32 & 30.17 & --     & --    \\
DriveDPO~\citep{shang2025drivedpo}          & --   & 62.02 & 30.62 & 166.80 & 26.79 \\
ThinkTwice~\citep{jia2023think}        & 0.95 & 62.44 & 31.23 &  69.33 & 16.22 \\
DriveTransformer~\citep{jia2025drivetransformer}  & 0.62 & 63.46 & 35.01 & 100.64 & 20.78 \\
DriveAdapter~\citep{jia2023driveadapter}      & 1.01 & 64.22 & 33.08 &  70.22 & 16.01 \\
RAP~\citep{feng2025rap}               & --   & 66.42 & 37.27 & 165.47 & 23.63 \\
ReCogDrive~\citep{li2025recogdrive}        & --   & 71.36 & 45.45 & 138.18 & 17.45 \\
DriveMoE~\citep{yang2025drivemoe}          & \textbf{0.38} & 74.22 & 48.64 & 175.96 & 15.31 \\
Orion~\citep{fu2025orion}             & 0.68 & 77.74 & \textbf{54.62} & 151.48 & 17.38 \\
UniDriveVLA~\citep{li2026unidrivevla}       & 0.72 & 78.37 & 51.82 & \textbf{198.86} & 11.78 \\
\midrule
VLGA (Ours) & 0.69 & \textbf{79.08} & 52.73 & 194.63 & 13.06 \\
\bottomrule
\end{tabular}
\end{table*}

% =================================================================
%  Table 4: Bench2Drive multi-ability breakdown
% =================================================================
\begin{table*}[t]
\centering
\caption{Per-skill closed-loop success rate (\%) on Bench2Drive.}
\label{tab:b2d_multi_ability}
\setlength{\tabcolsep}{6pt}
\small
\begin{tabular}{lcccccc}
\toprule
Method & Merging & Overtaking & Emergency Brake & Give Way & Traffic Sign & Mean $\uparrow$ \\
\midrule
AD-MLP~\citep{zhai2023rethinking}            &  0.00 &  0.00 &  0.00 &  0.00 &  4.35 &  0.87 \\
UniAD~\citep{hu2023planning}             & 14.10 & 17.78 & 21.67 & 10.00 & 14.21 & 15.55 \\
VAD~\citep{jiang2023vad}               &  8.11 & 24.44 & 18.64 & 20.00 & 19.15 & 18.07 \\
ThinkTwice~\citep{jia2023think}        & 13.72 & 22.93 & 52.99 & \textbf{50.00} & 47.78 & 37.48 \\
DriveAdapter~\citep{jia2023driveadapter}      & 14.55 & 22.61 & 54.04 & \textbf{50.00} & 50.45 & 38.33 \\
DriveTransformer~\citep{jia2025drivetransformer}  & 17.57 & 35.00 & 48.36 & 40.00 & 52.10 & 38.60 \\
ReCogDrive~\citep{li2025recogdrive}        & 29.73 & 20.00 & 69.09 & 20.00 & \textbf{71.34} & 42.03 \\
DriveMoE~\citep{yang2025drivemoe}          & 34.67 & 40.00 & 65.45 & 40.00 & 59.44 & 47.91 \\
Orion~\citep{fu2025orion}             & 25.00 & 71.11 & \textbf{78.33} & 33.00 & 69.15 & \textbf{54.72} \\
UniDriveVLA~\citep{li2026unidrivevla}       & \textbf{38.75} & \textbf{80.00} & 50.00 & 30.00 & 58.95 & 51.53 \\
\midrule
VLGA (Ours) & \textbf{38.75} & 77.78 & 55.00 & 40.00 & 53.68 & 53.04 \\
\bottomrule
\end{tabular}

\end{table*}

\subsection{Ablation Study}
\label{sec:ablation}

\textbf{Components of the geometric stream.} As shown in Tab.~\ref{tab:ablation}, 
both geometric components contribute monotonically, with dense
pointmap supervision driving the safety improvement. The geometry
expert alone gives a modest gain, and adding dense pointmap
reconstruction cuts the collision average to \textbf{0.136\%}, an
8.7\% relative reduction. This confirms that a geometric
stream is not sufficient: the dense reconstruction
objective is what keeps the stream task-relevant.

\begin{table}[t]
\centering
\caption{Ablation on VLGA's geometric stream components, on nuScenes
open-loop planning (ST-P3 metric, without ego status). Each row progressively
adds one component to the baseline.}
\label{tab:ablation}
\setlength{\tabcolsep}{8pt}
\small
\begin{tabular}{lcc}
\toprule
Configuration                          & L2 avg $\downarrow$ & Col avg $\downarrow$ \\
\midrule
Baseline (no geometry stream)          & 0.539               & 0.169\%              \\
+ Geometry expert                      & 0.529               & 0.149\%              \\
+ Pointmap aux supervision             & \textbf{0.524}      & \textbf{0.136\%}     \\
\bottomrule
\end{tabular}
\end{table}

\subsection{Analysis}
\label{sec:analysis}

\textbf{Where does the geometric prior help most?}
Comparing the per-skill closed-loop breakdown
(Tab.~\ref{tab:b2d_multi_ability}) to the open-loop collision pattern
(Tab.~\ref{tab:nusc_planning}), VLGA's gains concentrate on scenarios
and metrics that demand precise spatial reasoning. On Give Way and
Emergency Brake, where the policy must pause at the right offset or
brake at the right distance, VLGA improves over the strongest prior
VLA and matches it on Merging; on Overtaking and Traffic Sign, where
the bottleneck is reactive control and visual semantics rather than
dense geometry, the two are comparable. The same shape appears
open-loop, where VLGA's largest gains are on long-horizon collision
rather than mean L2. The geometry stream sharpens exactly the dense
spatial understanding tight-clearance driving needs, leaving the rest
of the policy largely unchanged.

\begin{figure*}[t]
\centering
\includegraphics[width=0.95\linewidth]{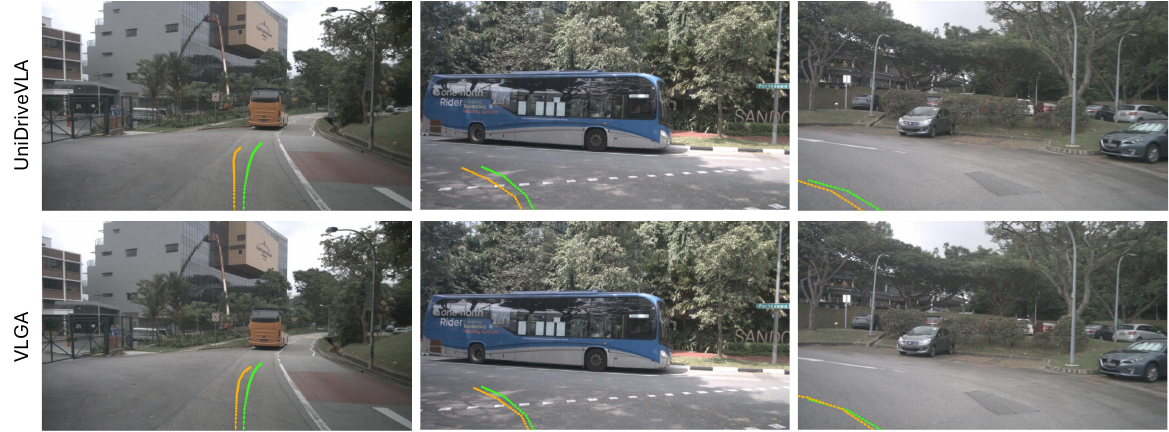}
\caption{\textbf{Qualitative planning comparison on the nuScenes validation
set.} The predicted 3-second trajectory (\textcolor[HTML]{FFB300}{yellow})
and ground truth (\textcolor[HTML]{39FF14}{green}) are projected onto the
front camera. VLGA stays closer to the ground truth through turns and around
nearby vehicles, while the baseline (UniDriveVLA) drifts laterally.}
\label{fig:qualitative}
\end{figure*}

\textbf{Qualitative planning comparison.}
Fig.~\ref{fig:qualitative} projects the predicted and ground-truth 3-second trajectories onto the front camera. VLGA stays closer to the ground truth through turns and around nearby vehicles, whereas the baseline drifts laterally. Since the two policies share the same vision-language and perception experts and differ only in VLGA's supervised geometry stream, this improved spatial grounding is attributable to the dense geometric representation the action expert conditions on.

% =================================================================
%  Figure 3: Qualitative planning comparison (ours vs UniDriveVLA)
% =================================================================

\section{Conclusion}
\label{sec:conclusion}

We introduced \textbf{VLGA}, a four-expert
vision-language-geometry-action policy that adds a dedicated,
parameter-isolated geometry expert and supervises it to reconstruct
the dense 3D world from LiDAR. This single addition concentrates gains
exactly where driving is hardest: VLGA records the lowest
collision rate among VLA methods on open-loop nuScenes and a
state-of-the-art driving score on closed-loop Bench2Drive, with its largest margins on
the most spatially demanding cases. Together, these results suggest
that explicit dense geometric supervision is one path toward
VLA driving policies that are not merely descriptive of the 3D world
but operationally grounded in it.

\textbf{Limitations \& Future Work.} VLA
driving policies incur substantial inference cost from their large
vision-language backbones, which constrains deployment on edge
compute; distillation and quantization tailored to the driving
setting are a natural direction.
On the method side, our pointmap
supervision is applied per-frame; extending it with temporal
consistency across the multi-frame input window is a natural
direction that may further sharpen long-horizon geometric reasoning.

\bibliographystyle{plainnat}
\bibliography{main}

\clearpage
\appendix

\section{Architectural Details}
\label{app:arch}

VLGA-Base and VLGA-Large use the public Qwen3-VL-2B and Qwen3-VL-8B
vision-language backbones with hidden dimensions $d_\text{VLM}$ of
$2048$ and $4096$, respectively; all new components introduced in this
paper are sized to match $d_\text{VLM}$.

\textbf{Geometry projector.} DVGT-2 emits per-patch features of
dimension $3072$. The per-patch projector
$f_\text{proj}\!:\mathbb{R}^{3072}\!\to\!\mathbb{R}^{d_\text{VLM}}$
is a 2-layer MLP that first applies a stride-2 spatial unshuffle on
the patch grid and then linearly projects to $d_\text{VLM}$.

\textbf{Pointmap decoder and head.} The five-layer transformer decoder
$\mathcal{D}$ has hidden dimension $d_\text{VLM}$, 8 attention heads,
a feed-forward expansion ratio of 4, and no dropout. Its output passes
through a 2-layer MLP head (hidden width $256$) followed by a
stride-$16$ pixel-shuffle (the DVGT-2 patch size), producing four
channels per pixel: the predicted $\hat{\mathbf{x}}_p$ and the
confidence logit $c_p$. The decoder and head are discarded at
inference.

\textbf{Action conditioning.} The action expert plans a 2D BEV
trajectory over six future steps. Three auxiliary inputs are projected
to $d_\text{VLM}$ and summed before they condition the expert: a
2-layer MLP encodes the ego status, comprising linear velocity, linear
acceleration, and a 3-class one-hot driving command; a lightweight MLP
encodes the past four trajectory steps; a 2-layer time MLP encodes the
flow-matching timestep.

\section{Training Hyperparameters}
\label{app:train}

Both training stages use AdamW with weight decay $10^{-7}$ and an
effective batch size of $128$ across $8$ H100 GPUs. EMA is applied to
all trainable parameters with momentum $2\times10^{-4}$ and a $2000$-iteration warmup. We enable non-reentrant gradient checkpointing on
the vision-language backbone, perception expert, geometry expert, and
action expert to fit the four-expert MoT within $80$\,GB of GPU
memory. Stage-specific settings are summarized in
Tab.~\ref{tab:supp_train}.

\begin{table}[h]
\centering
\small
\caption{Stage-specific training settings. Stage 1 (geometry) trains
only the new geometric components; Stage 2 (joint) additionally
unfreezes the action expert.}
\label{tab:supp_train}
\setlength{\tabcolsep}{8pt}
\begin{tabular}{lcc}
\toprule
                                & Stage 1 (geometry) & Stage 2 (joint) \\
\midrule
Base learning rate              & $1\times10^{-4}$ & $5\times10^{-5}$ \\
Epochs (nuScenes / Bench2Drive) & 10 / 3 & 30 / 7 \\
$\lambda_\text{pmap}$           & 1.0 & 0.1 \\
Action expert                   & frozen & unfrozen, flow-matching loss \\
\bottomrule
\end{tabular}
\end{table}

\section{Geometry Stream Visualization}
\label{app:point_traj}

\begin{figure}[h]
\centering
\includegraphics[width=0.82\linewidth]{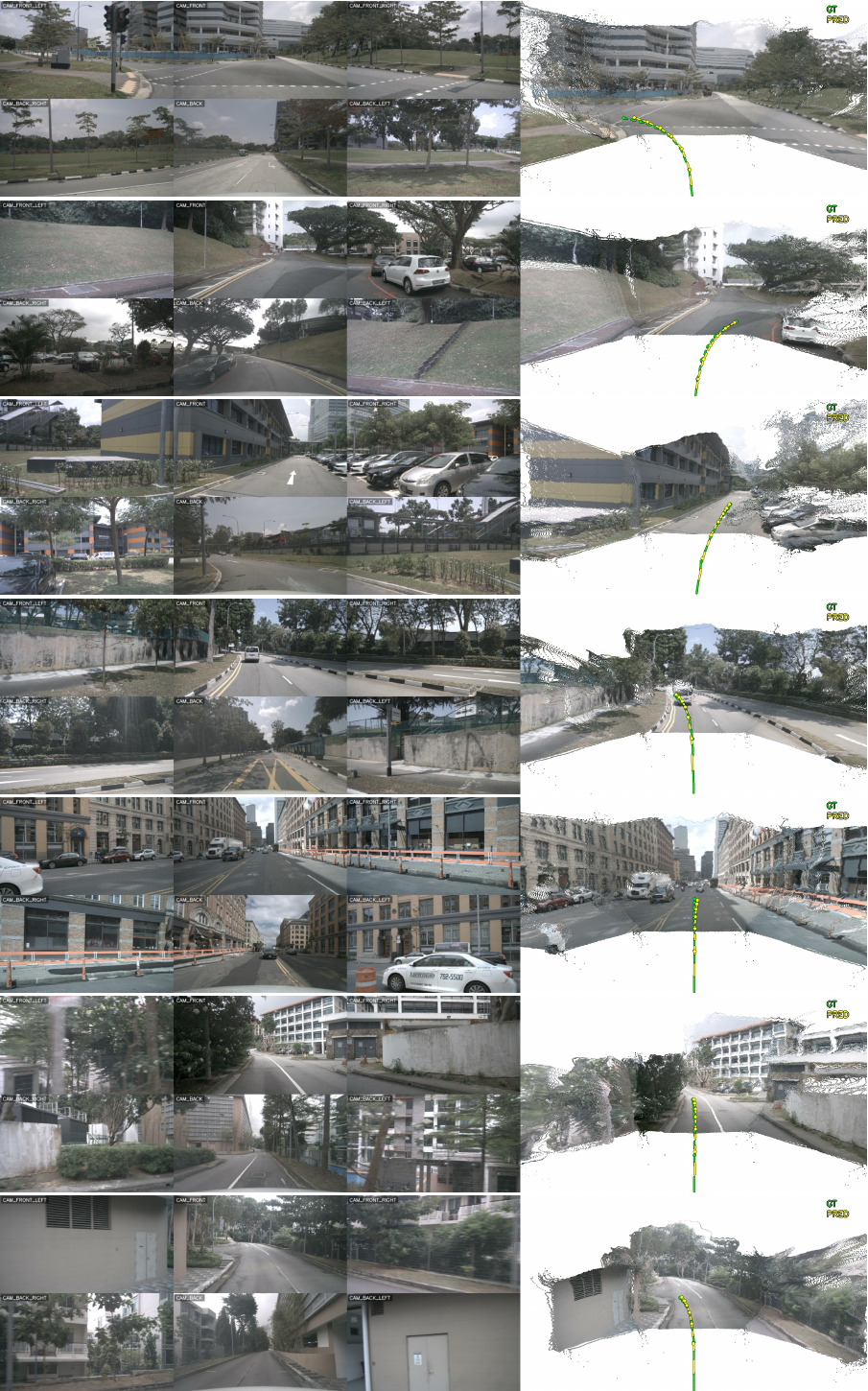}
\caption{\textbf{Pointmap reconstructions with predicted and
ground-truth trajectories on nuScenes.} Each row is one validation
sample. Left: the six surround-camera inputs. Right: the dense 3D
pointmap predicted by VLGA's geometry expert (decoded by
$\mathcal{D}$ for visualization), with the
\textcolor[HTML]{39FF14}{green} ground-truth and
\textcolor[HTML]{FFB300}{yellow} predicted ego trajectories overlaid.}
\label{fig:supp_point_traj}
\end{figure}

Fig.~\ref{fig:supp_point_traj} visualizes what the geometry expert
learns to encode. For seven nuScenes validation samples, the left grid
shows the six surround-camera inputs and the right panel shows the
dense 3D pointmap reconstructed from the geometry stream by the
training-time pointmap decoder $\mathcal{D}$. The ground-truth ego
trajectory (\textcolor[HTML]{39FF14}{green}) and VLGA's predicted
trajectory (\textcolor[HTML]{FFB300}{yellow}) are overlaid in the
reconstructed scene over the planning horizon. The reconstructions
capture road surface, lane structure, and surrounding obstacle
geometry, and the predicted trajectory closely tracks the ground
truth through turns and around static objects.

\end{document}